\documentclass{article}
\usepackage[accepted]{icml2020}
\pdfoutput=1
\usepackage[utf8]{inputenc} % allow utf-8 input
\usepackage[T1]{fontenc}    % use 8-bit T1 fonts
\usepackage{hyperref}       % hyperlinks
\hypersetup{hidelinks}
\usepackage{url}            % simple URL typesetting
\usepackage{booktabs}       % professional-quality tables
\usepackage{amsfonts}       % blackboard math symbols
\usepackage{nicefrac}       % compact symbols for 1/2, etc.
\usepackage{microtype}      % microtypography

\usepackage{enumitem}
\setenumerate{leftmargin=4mm}
\setitemize{leftmargin=4mm}

\title{Stochastic Differential Equations with Variational Wishart Diffusions}

\usepackage{microtype}
\usepackage{graphicx}
\usepackage{subfigure}
\usepackage{booktabs} % for professional tables
\usepackage[table,xcdraw]{xcolor}
\usepackage{multirow}
\usepackage{todonotes}
\usepackage[utf8]{inputenc}
\usepackage[english]{babel}
\usepackage[T1]{fontenc}
\usepackage{amsmath,amssymb}
\usepackage{bm}
\usepackage{amsthm}
\usepackage{mathtools}
\usepackage{pdflscape}
\usepackage{dsfont, mathrsfs}
\usepackage{mathrsfs}
\usepackage{wrapfig}
\graphicspath{{plots/}}
\newtheorem{definition}{Definition}

\renewcommand{\mathbf}{\boldsymbol} % this ensures that Greek alphabets are also bold (it will slant, though)

\renewcommand{\nolimits}{} % takes out the nolimits in equations

\usepackage{tikz}
\usetikzlibrary{arrows.meta}

\icmltitlerunning{Stochastic Differential Equations with Variational Wishart Diffusions}

\begin{document}

\twocolumn[
\icmltitle{Stochastic Differential Equations with Variational Wishart Diffusions}




\begin{icmlauthorlist}
\icmlauthor{Martin Jørgensen}{dtu}
\icmlauthor{Marc Peter Deisenroth}{ucl}
\icmlauthor{Hugh Salimbeni}{g}
\end{icmlauthorlist}

\icmlaffiliation{dtu}{Department for Mathematics and Computer Science, Technical University of Denmark}
\icmlaffiliation{ucl}{Department of Computer Science, University College London}
\icmlaffiliation{g}{G-Research}

\icmlcorrespondingauthor{Martin Jørgensen}{\href{mailto:marjor@dtu.dk}{marjor@dtu.dk}}


\vskip 0.3in
]



\printAffiliationsAndNotice{}  % leave blank if no need to mention equal contribution

\begin{abstract}
We present a Bayesian non-parametric way of inferring stochastic differential equations for both regression tasks and continuous-time dynamical modelling. The work has high emphasis on the \emph{stochastic} part of the differential equation, also known as the diffusion, and modelling it by means of Wishart processes. Further, we present a semi-parametric approach that allows the framework to scale to high dimensions. This successfully lead us onto how to model both latent and auto-regressive temporal systems with conditional heteroskedastic noise. We provide experimental evidence that modelling diffusion often improves performance and that this randomness in the differential equation can be essential to avoid overfitting.
\end{abstract}
\section{Introduction}

An endeared assumption to make when modelling multivariate phenomena with Gaussian processes (GPs) is that of independence between processes, i.e. every dimension of a multivariate phenomenon is modelled independently. Consider the case of a two-dimensional temporal process $\mathbf{x}_t$ evolving as
\begin{equation}
\label{simplemodel}
	\mathbf{x}_t:=f(\mathbf{x}_{t-1}) + \mathbf{\epsilon}_t,
\end{equation}
where $f(\mathbf{x}_{t-1})=(f_1(\mathbf{x}_{t-1}),f_2(\mathbf{x}_{t-1}))^\top$, $f_1$ and $f_2$ independent, and $\mathbf{\epsilon}_t\sim\mathcal{N}(\mathbf 0,\sigma^2\mathbf{I})$. This model is commonly used in the machine learning community and is easy to use and understand, but for many real-world cases the noise is too simplistic. In this paper, we will investigate the noise term $\mathbf{\epsilon}_t$ and also make it dependent on the state $\mathbf{x}_{t-1}$. This is also known as heteroskedastic noise. We will refer to the sequence of $\mathbf\epsilon_t$ as the \emph{diffusion} or \emph{process noise}.

\emph{Why} model the process noise? Assume that in the  example above, the two states represent meteorological measurements: rainfall and wind speed. Both are influenced by confounders, such as atmospheric pressure, which are not measured directly. This effect can in the case of the model in \eqref{simplemodel} only be modelled through the diffusion $\mathbf{\epsilon}$. Moreover, wind and rain may not correlate identically for all states of the confounders. 

%This type of dynamic modelling is central to fields like econometrics, climate science, and meteorology, emphasised with the \textit{AutoRegressive Conditional Heteroskedasticity} (ARCH) model \citep{engle1982autoregressive}.
Dynamical modelling with focus in the noise-term is not a new area of research. The most prominent one is the \textit{Auto-Regressive Conditional Heteroskedasticity} (ARCH) model \citep{engle1982autoregressive}, which is central to scientific fields like econometrics, climate science and meteorology.
The approach in these models is to estimate large process noise when the system is exposed to a shock, i.e. an unforeseen significant change in states. Thus, it does not depend on the value of some state, but rather on a linear combination of previous states.

In this paper, we address this shortcoming and introduce a model to handle the process noise by the use of \emph{Wishart processes}. Through this, we can sample covariance matrices dependent on the input state. This allows the system to evolve as a homogeneous system rather than independent sequences. By doing so, we can avoid propagating too much noise---which can often be the case with diagonal covariances---and potentially improve on modelling longer-range dependencies. Volatility modelling with GPs has been considered by \citet{gp_vol_wu, GWP, wilk_wish}.

For regression tasks, our model is closely related to several recent works exploring \emph{continuous-time} deep neural networks \citep{E2017, Haber2017, chen2018neural}. Here the notion of depth is no longer a discrete quantity (i.e. the number of hidden layers), but an interval on which a continuous flow is defined. In this view, continuous-time learning takes residual networks \citep{he2016deep} to their infinite limit, while remaining computationally feasible. The flow, parameterized by a differential equation, allows for time-series modelling, even with temporal observations that are not equidistant.

This line of work has been extended with stochastic equivalents \citep{twomey, tzen, liu, li2020scalable}, and the work by \mbox{\citet{look}}, who model the drift and diffusion of an SDE with Bayesian neural networks. These approaches make the framework more robust, as the original approach can fail even on simple tasks \citep{dupont}.

The work that inspired our model most was by \citet{hegde2019deep}. They model the random field that defines the SDE with a Gaussian field. They consider regression and classification problems. To this end, they can take deep GPs \citep{damianou2013deep, salimbeni2017doubly}  to their `infinite limit' while avoiding their degeneracy discussed by \citet{duvenaud2014avoiding}.

\textbf{Our main focus} throughout this paper lies on the \emph{stochasticity} of the flow, what impact it has and to which degree it can be tamed or manipulated to improve overall performance. Contributions:
\begin{itemize}
\item A model that unifies theory from conditional hetero-skedastic dynamics, stochastic differential equations (SDEs) and regression. We show how to perform variational inference in this model.
\item A scalable approach to extend the methods to high dimensional input without compromising with inter-dimensional independence assumptions.
\end{itemize}

\section{Background}\label{sec:background}
In this section, we give an overview of the relevant material on GPs, Wishart processes, and SDEs.
\subsection{Gaussian Processes}

A \emph{Gaussian process} (GP) is a distribution over functions $f: \mathbb{R}^d\rightarrow\mathbb{R}^D$, satisfying that for any finite set of points $\mathbf{X}:=\big(\mathbf{x}_1,\ldots,\mathbf{x}_N\big)^\top\in\mathbb{R}^{N\times d}$, the outputs $\big(f(\mathbf{x}_1),\ldots,f(\mathbf{x}_N)\big)^\top\in\mathbb{R}^{N\times D}$ are jointly Gaussian distributed. A GP is fully determined by a mean function $\mu:\mathbb{R}^d \rightarrow\mathbb{R}^D$ and a covariance function $c: \mathbb{R}^d\times\mathbb{R}^d\rightarrow \mathbb{R}^{D\times D}$. This notation is slightly unorthodox, and we will elaborate.

The usual convention when dealing with \emph{multi-output} GPs (i.e. $D>1$) is to assume $D$ i.i.d. processes that share the same covariance function \citep{alvarez2011}, which equivalently can be done by choosing the covariance matrix $\mathbf{K}=k(\mathbf{X},\mathbf{X})\otimes \mathbf{I}_D$, where $\otimes$ denotes the Kronecker product and $k$ is a covariance function for univariate output. For ease of notation we shall use $k^D(\mathbf{a},\mathbf{b}):=k(\mathbf{a},\mathbf{b})\otimes \mathbf{I}_D$; that is, $k(\mathbf{a},\mathbf{b})$ returns a kernel matrix of dimension number of rows in $\mathbf{a}$ times the number of rows in $\mathbf{b}$. This corresponds to the assumption of independence between output dimensions.
Furthermore, we write $\mathbf{f}:=f(\mathbf{X})$, $\mathbf{\mu}:=\text{vec}(\mu(\mathbf{X}))$ and denote by $\mathbf{K}$ the $ND\times ND$-matrix with ${K}_{i,j}=k^D(\mathbf{x}_i,\mathbf{x}_j)$. Then we can write in short $p(\mathbf{f})=\mathcal{N}(\mathbf{\mu},\mathbf{K})$.

As the number $N$ of training data points gets large, the size of $\mathbf{K}$ becomes a challenge as well, due to a required inversion during training/prediction. To circumvent this, we consider \emph{sparse} (or low-rank) GP methods. In this respect, we choose  $M$ auxiliary \emph{inducing} locations $\mathbf{Z}=\big(\mathbf{z}_1,\ldots,\mathbf{z}_M\big)^\top\in\mathbb{R}^{M\times d}$, and define their function values $\mathbf{u}:=f(\mathbf{Z})\in\mathbb{R}^{M\times D}$. Since any finite set of function values are jointly Gaussian, $p(\mathbf{f},\mathbf{u})$ is Gaussian as well, and we can write $p(\mathbf{f},\mathbf{u})=p(\mathbf{f}|\mathbf{u})p(\mathbf{u}),$ where $p(\mathbf{f}|\mathbf{u})=\mathcal{N}(\tilde{\mathbf{\mu}},\tilde{\mathbf{K}})$ with
\begin{align}
\tilde{\mathbf{\mu}}&=\mathbf{\mu}+\mathbf{\alpha}^\top\text{vec}(\mathbf{u}-\mu(\mathbf{Z})), \label{condmeanf}\\
\tilde{\mathbf{K}}&= k^D(\mathbf{X},\mathbf{X})+\mathbf{\alpha}^\top k^D(\mathbf{Z},\mathbf{Z})\mathbf{\alpha},\label{condvarf}
\end{align}
where $\mathbf{\alpha}=k^D(\mathbf{X,Z})k^D(\mathbf{Z},\mathbf{Z})^{-1}$. 
Here it becomes evident why this is computationally attractive, as we only have to deal with the inversion of $k^D(\mathbf{Z},\mathbf{Z})$, which due to the structure, only requires inversion of $k(\mathbf{Z},\mathbf{Z})$ of size $M\!\times\! M$. This is opposed to a matrix of size $ND\times ND$ had we not used the low-rank approximation and independence of GPs.

We will consider variational inference to marginalise $\mathbf{u}$ \citep{titsias}. Throughout the paper, we will choose our variational posterior to be
$q(\mathbf{f},\mathbf{u})=p(\mathbf{f}|\mathbf{u})q(\mathbf{u})$, where $ q(\mathbf u):=\mathcal{N}(\mathbf{m},\mathbf{S})$,
similar to \citet{bigdata-hensman}. Further, $q$ factorises over the dimensions, i.e. $q(\mathbf{u})=\prod_{j=1}^D\mathcal{N}(\mathbf{m}_j,\mathbf{S}_j)$, where $\mathbf{m}=(\mathbf{m}_1,\ldots,\mathbf{m}_D)$ and $\mathbf S$ is a block-diagonal $MD\!\times\!MD$-matrix, with block-diagonal entries $\{\mathbf{S}_j\}_{j=1}^D$. 
In this case, we can analytically marginalise $\mathbf{u}$ in \eqref{condmeanf} to obtain
\begin{align}
\label{vardist}
q(\mathbf{f})&=\int p(\mathbf{f}|\mathbf{u})q(\mathbf{u})d\mathbf{u}=\mathcal{N}(\mathbf{\mu}_{f}^q,\mathbf{K}_{f}^q),\\
\mathbf{\mu}_{f}^q&=\mathbf{\mu}+\mathbf{\alpha}^\top\text{vec}(\mathbf{m}-\mu(\mathbf{Z})), \\
%\quad\text{and} \quad
\mathbf{K}_{f}^q&= k^D(\mathbf{X},\mathbf{X})+\mathbf{\alpha}^\top \big(k^D(\mathbf{Z},\mathbf{Z})-\mathbf{S}\big)\mathbf{\alpha},
\end{align}
which resembles \eqref{condmeanf}--\eqref{condvarf}, but which is analytically tractable given variational parameters $\big\{\mathbf{m},\mathbf{S},\mathbf{Z}\big\}$.

Recall that a \emph{vector field} is a mapping $f:\mathbb{R}^d\rightarrow\mathbb{R}^D$ that associates a point in $\mathbb{R}^d$ with a vector in $\mathbb{R}^D$. A \emph{Gaussian (random) field} is a vector field, such that for any finite collection of points $\{\mathbf{x}_i\}_{i=1}^N$, their associated vectors in $\mathbb{R}^D$ are jointly Gaussian distributed, i.e. a Gaussian field \textit{is} a GP. We shall use both terminologies, but when we refer to a Gaussian field, we will think of the outputs as having a \emph{direction}.

\subsection{Wishart Processes}
The \emph{Wishart distribution} is a distribution over symmetric, positive semi-definite  matrices. It is the multidimensional generalisation of the $\chi^2$-distribution. Suppose $\mathbf{F}_v$ is a $D$-variate Gaussian vector for each $v=1,\ldots,\nu$ independently, say $\mathbf{F}_v\sim\mathcal{N}(\mathbf{0},\mathbf{A})$. Then $\mathbf{\Sigma}=\sum_{v=1}^{\nu}\mathbf{F}_v\mathbf{F}_v^\top$ is Wishart distributed with $\nu$ degrees of freedom and scale matrix $\mathbf{A}$. We write for short $\mathbf{\Sigma}\sim\mathcal{W}_D(\mathbf{A},\nu)$. By Bartlett's decomposition \citep{kshirsagar1959bartlett}, this can also be represented as $\mathbf{\Sigma} = \mathbf{L}\mathbf{F}\mathbf{F}^\top \mathbf{L}^\top$, where $\mathbf{F}$ is a $D\times\nu$-matrix with all entries unit Gaussian and $\mathbf{A}=\mathbf{LL}^\top$.

With this parametrization we  define Wishart processes, as in \cite{GWP}:
\begin{definition}
	Let $\mathbf{L}$ be a $D\times D$ matrix, such that $\mathbf{LL}^\top$ is positive semidefinite and $f_{d,v}\sim \mathcal{GP}\big(0,k_{d,v}(\mathbf{x},\mathbf{x}')\big)$ independently for every $d=1,\ldots,D$ and $v=1\ldots,\nu$, where $\nu\geq D$. Then if 
	\begin{equation}
	\Sigma(\mathbf{x}) = \mathbf{L}\Bigg(\sum_{v=1}^\nu \mathbf{f}_v(\mathbf{x})\mathbf{f}_v^\top(\mathbf{x})\Bigg)\mathbf{L}^\top
	\end{equation}
	is Wishart distributed for any marginal $\mathbf{x}$, and if for any finite collection of points $\mathbf{X}=\{\mathbf{x}_i\}_{i=1}^N$ the joint distribution $\Sigma(\mathbf{X})$ is determined through the covariance functions $k_{d,v}$, then $\Sigma(\cdot)$ is a Wishart process. We will write
	\begin{equation}
	\Sigma \sim \mathcal{WP}_D(\mathbf{LL}^\top,\nu,\kappa),
	\end{equation}
	where $\kappa$ is the collection of covariance functions $\{k_{d,v}\}$.
\end{definition}
If $\mathbf{\Sigma}$ follows a Wishart distribution with $\nu$ degrees of freedom and scale matrix $\mathbf{LL}^\top$ of size $D\times D$, then for some $\rho\times D$-matrix $\mathbf{R}$ of rank $\rho$, we have that $\mathbf{R\Sigma R}^\top\sim \mathcal{W}_\rho(\mathbf{RLL}^\top\mathbf{R}^\top,\nu)$. That is, $\mathbf{R\Sigma R}^\top$ is Wishart distributed on the space of $\rho\times\rho$ symmetric, positive semi-definite matrices.

The Wishart distribution is closely related to the Gaussian distribution in a Bayesian framework, as it is the conjugate prior to the precision matrix of a multivariate Gaussian. Furthermore, it is the distribution of the maximum likelihood estimator of the covariance matrix.

The Wishart process is a slight misnomer as the posterior processes are \emph{not} marginally Wishart. This is due to the mean function not being constant $0$, and a more accurate name could be \emph{matrix-Gamma} processes. We shall not refrain from the usual terminology: a Wishart process is a stochastic process, whose \emph{prior} is a Wishart process.
\subsection{Stochastic Differential Equations} 
We will consider SDEs of the form
\begin{equation}\label{sde}
d\mathbf{x}_t = \mu(\mathbf{x}_t)dt + \sqrt{\mathbf{\Sigma}(\mathbf{x}_t)}dB_t,
\end{equation}
where the last term of the right-hand side is the Itô integral \citep{ito}. The solution $\mathbf{x}_t$ is a stochastic process, often referred to as a \emph{diffusion process}, and $\mu$ and $\mathbf{\Sigma}$ are the drift and diffusion coefficients, respectively. In (\ref{sde}), $B_t$ denotes the Brownian motion. 

The Brownian motion is the GP satisfying that all increments are independent in the sense that, for $0\leq s_1< t_1\leq s_2< t_2$, then $B_{t_1-s_1}$ is independent from $B_{t_2-s_2}$. Further, any increment has distribution $B_t-B_s\sim \mathcal{N}(0,t-s)$. Lastly, $B_0=0$. This is equivalent to the GP with constant mean function $0$ and covariance function $(t, s)\mapsto \min\{s,t\}$ \citep{rasmussen:book}.

Given some initial condition (e.g. $\mathbf{x}_0=\mathbf{0}$), we can generate \emph{sample paths} $[0,T]\to \mathbb{R}^D$ by the Euler-Maruyama method. Euler-Maruyama \citep{kloeden2013numerical} finely discretizes the temporal dimension $0=t_0<t_1<\ldots<t_l=T$, and \emph{pushes} $\mathbf{x}_{t_i}$ along the vector field
$
\mathbf{x}_{t_{i+1}}=\mathbf{x}_{t_{i}}+\mathbf{\mu}(\mathbf{x}_{t_i})\Delta_i+\sqrt{\mathbf{\Sigma}(\mathbf{x}_{t_i})\Delta_i}\mathbf{N}
$,
where $\mathbf{N}\sim \mathcal{N}(\mathbf{0},\mathbf{I}_D)$ and $\Delta_i = t_{i+1}-t_i$.

\section{Model and variational inference}\label{sec:model}
We consider a random field $f:\mathbb{R}^D\times[0,T]\rightarrow \mathbb{R}^D$ and a GP $g:\mathbb{R}^D\rightarrow \mathbb{R}^\eta$. Their priors are
\begin{equation}
f\sim \mathcal{GP}(0,k_f(\cdot,\cdot)\otimes \mathbf{I}_D), \quad g\sim \mathcal{GP}(0,k_g(\cdot,\cdot)\otimes \mathbf{I}_\eta).
\end{equation} 
We also have a Wishart process $\Sigma:\mathbb{R}^D\times[0,T]\rightarrow \mathcal{G}$, where $\mathcal{G}$ is the set of symmetric, positive semi-definite $D\times D$ matrices; the specific prior on this will follow in Section \ref{sec:wgp}.
We will approximate the posteriors of $f$, $g$ and $\Sigma$ with variational inference, but first we will formalise the model.

\begin{figure*}[t]
	\centering
	\hfill
    \subfigure[Graphical model based on Eq.~\eqref{factoredmodel}]{
		\begin{tikzpicture}[scale=0.8]
		\begin{scope}[every node/.style={circle,thick,draw,minimum size=0.9cm}]
		\node (x0) at (0,0) {$\mathbf{x}_0$};
		\node[draw=blue!80] (xt) at (2,0)	{$\mathbf{x}_t$};
		\node (xT) at (4,0) {$\mathbf{x}_T$};
		\node[draw=blue!80] (Sigma) at (1,2) {$\mathbf{\Sigma}_t$};
		\node[draw=blue!80] (ft) at (3,2) {$\mathbf{f}_t$};
		\node (g) at (6,0) {$\mathbf{g}$};
		\node (y) at (8,0) {$\mathbf{y}$};
		\end{scope}
		\begin{scope}[every node/.style={rectangle,thick,draw, minimum size=0.7cm}]
		\node (usig) at (-0.5,2) {$\mathbf{u}_{\Sigma}$};
		\node (uf) at (4.5,2) {$\mathbf{u}_f$};
		\node (ug) at (6,1.5) {$\mathbf{u}_g$};
		\end{scope}
		\begin{scope}[>={Stealth[black]},
		every edge/.style={draw,very thick},
		every node/.style={circle}]
		\path [->] (x0) edge (xt);
		\path[draw=blue!80] [->] (xt) edge[bend left=40] (Sigma);
		\path[draw=blue!80] [->] (Sigma) edge[bend left=40] (ft);
		\path[draw=blue!80] [->] (ft) edge[bend left=40] (xt);
		\path [->] (xt) edge (xT);
		\path [->] (xT) edge (g);
		\path [->] (usig) edge (Sigma);
		\path [->] (uf) edge (ft);
		\path [->] (ug) edge (g);
		\path [->] (g) edge (y);
		\end{scope}
		\end{tikzpicture} 
		\label{fig:left}
		}
\hfill
\subfigure[Cycle from \subref{fig:left} and how it moves along the time-axis.]{
		\begin{tikzpicture}[scale=0.7]
		\begin{scope}[every node/.style={}]
		\node (dots1) at (0,0) {$\mathbf{\cdots}$};
		\node (dots2) at (10,1) {$\cdots$};
		\node (dots3) at (10,-1) {$\cdots$};
		\node (xsis) at (7,-1) {$\mathbf{x}_s :=\mathbf{f}_t$};
		\node (ftis) at (8.2,1.5) {$\mathbf{f}_t=\mu(\mathbf{x}_t)(s-t)+\sqrt{\Sigma(\mathbf{x}_t)}\mathbf{N}$};
		\end{scope}
		
		\begin{scope}[every node/.style={circle,thick,draw,minimum size = 0.9cm}]
		\node[draw=blue!80] (xt) at (2,0) {$\mathbf{x}_t$};
		\node[draw=blue!80] (xs) at (8,0) {$\mathbf{x}_{s}$};
		\node[draw=blue!80] (muxt) at (4,1) {$\mu(\cdot)$};
		\node[draw=blue!80] (Sigxt) at (4,-1) {$\Sigma(\cdot)$};
		\node[draw=blue!80] (ft) at (6,0) {$\mathbf{f}_t$};
		\end{scope}
		
		\begin{scope}[>={Stealth[black]},
		every edge/.style={draw,very thick},
		every node/.style={circle}]
		\path[draw=blue!80] [->] (xt) edge (Sigxt);
		\path[draw=blue!80] [->] (xt) edge (muxt);
		\path[draw=blue!80] [->] (dots1) edge (xt);
		\path[draw=blue!80] [->] (muxt) edge (ft);
		\path[draw=blue!80] [->] (Sigxt) edge (ft);
		\path[draw=blue!80] [->] (ft) edge (xs);
		\path[draw=blue!80] [->] (xs) edge (dots3);
		\path[draw=blue!80] [->] (xs) edge (dots2);
		\end{scope}
		\end{tikzpicture}
		\label{fig:right}
}
	\caption{\subref{fig:left} Graphical model based on the factorisation in Eq.~\eqref{factoredmodel}; \subref{fig:right} The cycle from \subref{fig:left}, which represents the \emph{field} $f$, and how it moves along the time-axis. Here $\mathbf{N}\sim\mathcal{N}(\mathbf{0},(s-t)\mathbf{I})$. Blue represents the flow/SDE, square nodes are variational variables.}
	\label{graphicalmodel}
\end{figure*}
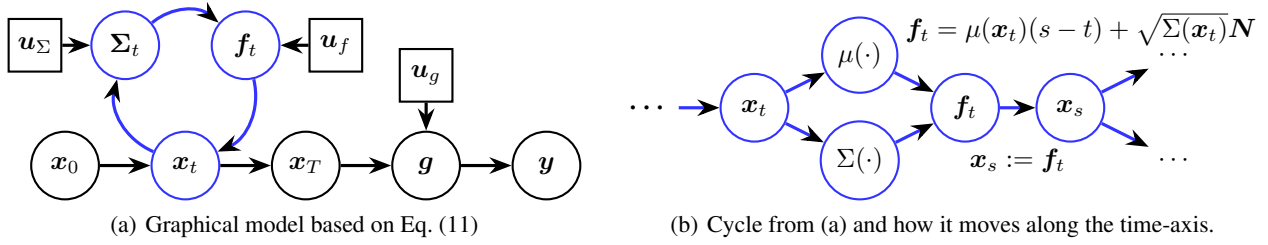
We propose a continuous-time deep learning model that can propagate noise in high-dimensions. This is done by letting the diffusion coefficient $\mathbf{\Sigma}(\mathbf{x}_t)$ of an SDE be governed by a Wishart process. The model we present factorises as
\begin{align}\label{factoredmodel}
\begin{split}
p(\mathbf{y},\Theta)=& p(\mathbf{y}|\mathbf{g})p(\mathbf{g}|\mathbf{x}_T,\mathbf{u}_g)p(\mathbf{u}_g)p(\mathbf{x}_T|\mathbf{f})\\ &\cdot p(\mathbf{f}|\mathbf{\Sigma},\mathbf{u}_f)p(\mathbf{u_f})p(\mathbf{\Sigma}|\mathbf{u}_\Sigma)p(\mathbf{u}_\Sigma),
\end{split}
\end{align}
where $\Theta:=\big\{ \mathbf{g},\mathbf{u}_g,\mathbf{x}_T, \mathbf{f},\mathbf{u}_f, \mathbf{\Sigma},\mathbf{u}_{\Sigma}\big\}$ denotes all variables to be marginalised. We assume that data $\mathcal{D}=\big\{(\mathbf{x}_i,\mathbf{y}_i)\big\}_{i=1}^N$ is i.i.d. given the process, such that $p(\mathbf{y}|\mathbf{g})=\prod_{i=1}^{N}p(\mathbf{y}_i|\mathbf{g}_i)$. We approximate the posterior of $g$ with the variational distribution as in (\ref{vardist}), i.e.
\begin{align}
q(\mathbf{g}_i)&=\int p(\mathbf{g}_i|\mathbf{u}_g)q(\mathbf{u}_g)d\mathbf{u}_g\\
&=\mathcal{N}(\tilde{\mu}_g(\mathbf{x}_i),\tilde{k}_g(\mathbf{x}_i,\mathbf{x}_i)),
\end{align}
where
\begin{align}
\tilde{\mu}_g(\mathbf{x}_i) &= \boldsymbol\alpha_g^\top(\mathbf{x}_i) \text{vec}(\mathbf{m}_{g}),\label{onlymarg1}\\
\tilde{k}_g(\mathbf{x}_i,\mathbf{x}_i)&=k_g^\eta(\mathbf{x}_i,\mathbf{x}_i)\label{onlymarg2}\\
&\quad +\boldsymbol\alpha_g^\top(\mathbf{x}_i)\big(k_g^\eta(\mathbf{Z}_g,\mathbf{Z}_g)-\mathbf{S}_g\big)\boldsymbol\alpha_g(\mathbf{x}_i),\nonumber
\end{align}
where $\boldsymbol\alpha_g(\mathbf{x}_i):=k_g^\eta(\mathbf{x}_i,\mathbf{Z}_g)k^\eta_g(\mathbf{Z}_g,\mathbf{Z}_g)^{-1}$. Here $\mathbf{m}_g$ is an $M\times\eta$ matrix, and $\mathbf{S}_g$ is an $M\eta\times M\eta$-matrix, constructed as $\eta$ different $M\times M$-matrices $\mathbf{S}_g=\{\mathbf{S}_j\}_j^\eta$.
During inference \citep{sparse-unifying}, we additionally assume that the marginals $g_i=g(\mathbf{x}_i)$ are independent when conditioned on $\mathbf{u}_g$. This is an approximation to make inference computationally easier. %Eqs.~(\ref{onlymarg1})--(\ref{onlymarg2}) reveal that the marginals $g_i$ depend only on the marginal $\mathbf{x}_i$ for any $i=1,\ldots,N$.

The inputs to $g$ are given as the state distribution of an SDE at a fixed time point $T\geq0$. We construct this SDE from the viewpoint of a random field. Consider the random walk with step size $\Delta$ on the simplest Gaussian field, where any state has mean $\mathbf{\mu}$ and covariance $\mathbf{\Sigma}$. For any time point $t$, the state distribution is tractable, i.e. $p(\mathbf{x}_t)=\mathbf{x}_0+\sum_{s=1}^S\mathcal{N}(\Delta_s\mathbf{\mu},\Delta_s\mathbf{\Sigma})$, where $\sum\Delta_s=t$ and $S$ is any positive integer.

For a state-dependent Gaussian field, we define the random walk
\begin{equation}
\mathbf{x}_{t+\Delta}=\mathbf{x}_t + \mu(\mathbf{x}_t)\Delta + \sqrt{\boldsymbol\Sigma(\mathbf{x}_t)\Delta}\mathbf{N}, 
\end{equation} 
with $\mathbf{N}\sim \mathcal{N}(\mathbf{0},\mathbf{I})$. Given an initial condition $\mathbf{x}_0$, the state $\mathbf{x}_S$  after $S$ steps is given by
\begin{equation}
\mathbf{x}_S=\mathbf{x}_0 + \sum\nolimits_{s=0}^{S-1}\Big(\mu(\mathbf{x}_s)\Delta + \sqrt{\boldsymbol\Sigma(\mathbf{x}_s)\Delta}\mathbf{N}\Big).
\end{equation}
In the limit $\Delta\rightarrow 0$, this random walk dynamical system is given by the diffusion process \citep{durrett}
\begin{equation}\label{SDE-field}
\mathbf{x}_T-\mathbf{x}_0=\int_{0}^{T}\mu(\mathbf{x}_t)dt + \int_{0}^T\sqrt{\boldsymbol\Sigma(\mathbf{x}_t)}dB_t,
\end{equation}
where $B$ is a Brownian motion. This is an SDE in the Îto-sense, which we numerically can solve by the Euler-Maruyama method. We will see that by a particular choice of variational distribution that $\mathbf{\Sigma}(\mathbf{x}_t)$ will be the realisation of a Wishart process. 
The coefficients in \eqref{SDE-field} are determined as the mean and covariance of a Gaussian field $f$. The posterior of $f$ is approximated with a Gaussian $q(\mathbf{f}_i)=\mathcal{N}(\mu_f^q(\mathbf{x}_i),k_f^q(\mathbf{x}_i,\mathbf{x}_i))$, where
\begin{align}
\mu_f^q(\mathbf{x}_i) =& \boldsymbol\alpha_f^\top(\mathbf{x}_i) \text{vec}(\mathbf{m}_{f}),\label{flow-mean}\\
k^q_f(\mathbf{x}_i,\mathbf{x}_i)=& k_f^D(\mathbf{x}_i,\mathbf{x}_i)\\
+&\boldsymbol\alpha_f^\top(\mathbf{x}_i)\big(k_f^D(\mathbf{Z}_f,\mathbf{Z}_f)-\mathbf{S}_f\big)\boldsymbol\alpha_f(\mathbf{x}_i),\label{flow-var}
\end{align}
and $\boldsymbol\alpha_{f}(\cdot)=k_f^D(\cdot,\mathbf{Z}_f)k^D_f(\mathbf{Z}_f,\mathbf{Z}_f)^{-1}$.

So far, we have seen how we move a data point $\mathbf{x}_0$ through the SDE \eqref{SDE-field} to $\mathbf{x}_T$, and further through the GP $g$, to make a prediction. However, each coordinate of $\mathbf{x}$ moves independently. By introducing the Wishart process, we will see how this assumption is removed.

\subsection{Wishart-priored Gaussian random field}\label{sec:wgp}
We are still considering the Gaussian field $f$, whose posterior is approximated by the variational distribution $q(\mathbf{f})$. To regularise (or learn) the noise propagated through this field into $g$, while remaining within the Bayesian variational framework, we define a hierarchical model as
\begin{align}
p(\mathbf{f})\!=\!\int p(\mathbf{f}|\mathbf{u}_f\!,\!\mathbf{\Sigma})p(\mathbf{u}_f)p(\mathbf{\Sigma}|\mathbf{u}_\Sigma)p(\mathbf{u}_\Sigma)d\{\mathbf{\Sigma}\!,\!\mathbf{u}_f\!,\!\mathbf{u}_\Sigma\},
\end{align}
where  ${\Sigma}$ is a Wishart process. Specifically, its prior is
\begin{equation}
\mathbf{\Sigma} \sim \mathcal{WP}_D(\mathbf{LL}^\top,\nu,k_f),
\end{equation}
that is any marginal $\mathbf{\Sigma}(\mathbf{x}_t)=\mathbf{L}\mathbf{JJ}^\top \mathbf{L}^\top$, where $\mathbf{J}$ is the $D\times\nu$-matrix with all independent entries $j_{d,v}(\mathbf{x}_t)$ drawn from GP's that share the same prior $j_{d,v}(\cdot)\sim \mathcal{GP}(0,k_f(\cdot,\cdot))$. To approximate the posterior of the Wishart process we choose a variational distribution 
\begin{equation}
q(\mathbf{J},\mathbf{u}_\Sigma) =q(\mathbf{J}|\mathbf{u}_\Sigma)q(\mathbf{u}_\Sigma):=p(\mathbf{J}|\mathbf{u}_\Sigma)q(\mathbf{u}_\Sigma),
\end{equation}
where $q(\mathbf{u}_\Sigma)=\prod_{d=1}^D\prod_{v=1}^\nu \mathcal{N}(\mathbf{m}_{d,v}^\Sigma,\mathbf{S}_{d,v}^\Sigma)$. Here,  $\mathbf{m}_{d,v}^\Sigma$ is $M\!\times\! 1$ and $\mathbf{S}_{d,v}^\Sigma$ is $M\!\times\! M$ for each pair $\{d,v\}$. Notice the same kernel is used for the Wishart process as is used for the random field $f$, that is: only one kernel \emph{controls} the vector field $f$. The posterior of $\boldsymbol\Sigma$ is naturally defined through the posterior of $J$. Given our choice of kernel, this approximate posterior is identical to Eqs.~\eqref{flow-mean}-\eqref{flow-var}, only changing the variational parameters to $\mathbf{m}_\Sigma$ and $\mathbf{S}_\Sigma$, and $D$ changes to $D\nu$. 

What remains to be defined in (\ref{factoredmodel}) is $p(\mathbf{f}|\mathbf{\Sigma},\mathbf{u}_f)$. Since $\mathbf{\Sigma}(\mathbf{x}_t)$ is a $D\!\times \!D$-matrix we define
\begin{align}
p\big(\mathbf{f}|\{\mathbf{\Sigma}(\mathbf{x}_i)\}_{i=1}^N,\mathbf{u}_f\big)&=\mathcal{N}\big(\tilde{\mu}(\mathbf{X}),\tilde{k}_f^\Sigma(\mathbf{X},\mathbf{X})\big),\\
\tilde{\mu}(\mathbf{x}_i)&=\boldsymbol\alpha_{f}^\top(\mathbf{x}_i)\text{vec}(\mathbf{u}_f),\\
\tilde{k}_{f}^\Sigma(\mathbf{x}_i,\mathbf{x}_j)&=\big(\mathbf{\Sigma}(\mathbf{x}_i)-\mathbf{h}_{ij}\big)\delta_{ij} + \mathbf{h}_{ij},
\end{align}
where $\mathbf{h}_{ij}=\boldsymbol\alpha_{f}(\mathbf{x}_i)^\top k_f^D(\mathbf{Z}_f,\mathbf{Z}_f)\alpha_f(\mathbf{x}_j)$ and $\delta_{ij}$ is Kronecker's delta. Notice this, conditioned on the Wishart process, constitutes a FITC-type model \citep{snelson2006sparse}.

This goes beyond the assumption of independent output dimensions, and instead makes the model learn the inter-dimensional dependence structure through the Wishart process $\Sigma$. This structure shall also be learned in the variational inference setup. The posterior of conditional $\mathbf{f}$ is approximated by 
\begin{align}
\begin{split}
q(\mathbf{f},\mathbf{u}_f|\{\mathbf{\Sigma}(\mathbf{x}_i)\}_{i=1}^N)&=q(\mathbf{f}|\{\mathbf{\Sigma}(\mathbf{x}_i)\}_{i=1}^N,\mathbf{u}_f)q(\mathbf{u}_f)\\&=p(\mathbf{f}|\{\mathbf{\Sigma}(\mathbf{x}_i)\}_{i=1}^N,\mathbf{u}_f)q(\mathbf{u}_f),
\end{split}
\end{align}
where $q(\mathbf{u}_f):=\mathcal{N}(\mathbf{m}_f,k_f^D(\mathbf{Z}_f,\mathbf{Z}_f))$. At first, this might seem restrictive, but covariance estimation is already in $\Sigma$ and the variational approximation is the simple expression
\begin{equation}
q(\mathbf{f}|\{\mathbf{\Sigma}(\mathbf{x}_i)\}_{i=1}^N)=\prod_{i=1}^{N}\mathcal{N}\big(\alpha_{f}^\top(\mathbf{x}_i)\mathbf{m}_f,\mathbf{\Sigma}(\mathbf{x}_i)\big).
\end{equation}
The marginalisation can then be computed with Jensen's inequality 
\begin{align}
\log p(\mathbf{y})=& \log \!\int\! p(\mathbf{y},\Theta)d\Theta\nonumber\\
\geq& \!\int\! \log \Big(\frac{p(\mathbf{y},\Theta)}{q(\Theta)}\Big)q(\Theta)d\Theta\nonumber\\
=&\!\int\!\log p(\mathbf{y}|\mathbf{g})q\big(\mathbf{g}|\Theta\!\setminus\!\{\mathbf{g}\}\big)d\Theta\\
& - \text{KL}\big(q(\mathbf{u}_g)\|p(\mathbf{u}_g)\big)\nonumber\\
& -\text{KL}\big(q(\mathbf{u}_f)\|p(\mathbf{u}_f)\big)\! -\! \text{KL}\big(q(\mathbf{u}_\Sigma)\|p(\mathbf{u}_\Sigma)\big),\nonumber
\end{align}
or, in a more straightforward language,
\begin{align}\label{ELBO}
\log p(\mathbf{y})&\geq \mathbb{E}_{q(g)}[\log p(\mathbf{y}|\mathbf{g})]-\text{KL}\big(q(\mathbf{u}_g)\|p(\mathbf{u}_g)\big)\\
&\quad -\text{KL}\big(q(\mathbf{u}_f)\|p(\mathbf{u}_f)\big)\! -\! \text{KL}\big(q(\mathbf{u}_\Sigma)\|p(\mathbf{u}_\Sigma)\big).\nonumber
\end{align}
The right-hand side in~\eqref{ELBO} is the so-called \emph{evidence lower bound} (ELBO). The first term, the expectation, is analytically intractable, due to $q(g)$ being non-conjugate to the likelihood. Therefore, we determine it numerically with Monte Carlo (MC) or with Gauss-Hermite quadrature \citep{hensman2015mcmc}. With MC, often a few samples are enough for reliable inference \citep{salimans2013fixed}.

The KL-terms in \eqref{ELBO} can be computed analytically as they all involve multivariate Gaussians. Still, due to some of the modelling constraints, it is helpful to write them out, which yields
\begin{align}
	&\text{KL}\big(q(\mathbf{u}_g)\|p(\mathbf{u}_g)\big)=\sum\nolimits_{d=1}^{\eta}\text{KL}\big(q(\mathbf{u}_{g_d})\|p(\mathbf{u}_{g_d})\big),\\
	&\text{KL}\big(q(\mathbf{u}_\Sigma)\|p(\mathbf{u}_\Sigma)\big) =\sum_{d=1}^{D}\sum_{v=1}^{\nu}\text{KL}\big(q(\mathbf{u}_{\Sigma_{d,v}})\|p(\mathbf{u}_{\Sigma_{d,v}})\big),
	\label{bigKL}
\end{align}
where in both instances we used the independence between the GPs. The remaining one is special. Since both distribution share the same covariance it reduces to
\begin{equation}
	\text{KL}\big(q(\mathbf{u}_f)\|p(\mathbf{u}_f)\big)=\frac{1}{2}\sum\nolimits_{d=1}^{D}\mathbf{m}_{f_d}^\top k_f^D(\mathbf{Z}_f,\mathbf{Z}_f)^{-1}\mathbf{m}_{f_d}.
\end{equation}
Here, $k_f^D(\mathbf{Z}_f,\mathbf{Z}_f)^{-1}$ is already known from the computation of \eqref{bigKL}, as the kernel and inducing locations are shared.

Summarising this section, we have inputs $\mathbf{x}_0:=\mathbf{x}$ that are warped through an SDE (governed by a random field $f$) with drift $\mu$ and diffusion $\Sigma$ that is driven by \emph{one} kernel $k_f^D$. The value of this SDE, at some given time $T$, is then used as input to a final layer $g$, i.e. $g(\mathbf{x}_T)$ predicts targets $y(\mathbf{x})$. All this is inferred by maximising the ELBO~\eqref{ELBO}.

\subsection{Complexity and scalability}\label{sec:complex}
The computational cost of estimating $\mathbf{\Sigma}$ with a Wishart, as opposed to a diagonal matrix, can be burdensome. For the diagonal, the cost is $\mathcal{O}(DNM^2)$ since we need to compute \eqref{condvarf} $D$ times. Sampling $D\nu$ GP values and then matrix-multiplying it with a $D\times\nu$ matrix is of complexity $\mathcal{O}(D\nu NM^2 + D\nu D)$. Hence, if we, for simplicity, let $\nu=D$, we have overhead cost of $\mathcal{O}(D^2NM^2 + D^3)$. Note this is only the computational budget associated with the diffusion coefficients of the random field; the most costly one.

On this inspection, we propose a way to overcome a too heavy burden if $D$ is large. Naturally this involves an approximation; this time a low-rank approximation on the dimensionality-axis. Recall that, if $\mathbf{\Sigma}_\rho\sim \mathcal{WP}_\rho(\mathbf{I},\nu,\kappa)$, then $\mathbf{\Sigma}_D:=\mathbf{L\Sigma}_\rho \mathbf{L}^\top\sim\mathcal{WP}_D(\mathbf{LL}^\top,\nu,\kappa)$. The matrices naturally are of rank $\rho\ll D$. The computational overhead is reduced to $\mathcal{O}(\rho^2NM^2+D\rho^2)$ if $\nu=\rho$. This same structure was introduced by \citet{wilk_wish} for time-series modelling of financial data; and it reminisces the structure of Semiparametric Latent Factor Models (SLFM) \citep{SLFM}.
%in the sense that they let $\rho$ GPs find a non-linear structure of a $D$-dimensional output, and leave the remainder to be predicted by a parametric $D\!\times\! \rho$-matrix. 
That is, we have $\rho$ GPs, and the $D$-dimensional outputs are all linear combinations of these.
For clarity, we need only to compute/sample $\sqrt{\mathbf{\Sigma}_D}=\mathbf{LJ}$, where $\mathbf{J}$ is a $\rho\times\nu$ matrix, with GP values according to the approximate posterior $q(\mathbf{J})$, where $D$ replaced by $\rho$. 

\subsection{Further model specifications}
If $\rho$ is too small it can be difficult to identify a good diffusion coefficient as the matrix is too restricted by the low rank. One possible way to overcome this is too add `white noise' to the matrix
\begin{equation}
	\mathbf{\Sigma}=\mathbf{LFF}^\top\mathbf{L}^\top + \mathbf{\Lambda},
\end{equation}
where $\mathbf{\Lambda}$ is a diagonal $D\! \times\! D$-matrix. In many situations, this will ensure that the diffusion is full rank, and this provides more freedom in estimating the marginal variances. However, if the values on the diagonal of $\mathbf{\Lambda}$ are estimated by maximum likelihood, we have to be cautious. If $\mathbf{\Lambda}$ becomes to `dominant', inference can turn off the Wishart-part, potentially leading to overfitting.

Consider the matrix $\mathbf{L}$, that makes up the scale matrix of the Wishart process. It is fully inferred by maximum likelihood, hence there is no KL-term to regularise it. Effectively, this can turn off the stochasticity of the flow by making some matrix norm of $\mathbf{L}$ be approximately zero. Then the flow is only determined by its drift and overfitting is a likely scenario.

To alleviate this concern we propose to regularise $\mathbf{L}$ by its rownorms. That is,
\begin{equation}
	\forall d=1,\ldots,D:\quad \sum_{r=1}^{\rho}L_{d,r}^2 = 1,
\end{equation}
where $L_{d,r}$ denotes the entries of $\mathbf{L}$. First of all, this ensures that the prior variance for all dimensions is determined by the kernel hyperparameters, as it makes the diagonal of the scale matrix $\mathbf{LL}^\top$ equal to $1$. This way the variance in each dimension is a `fair' linear combination of the $\rho$ GPs that control the Wishart.

\subsection{Extending to time series}
The specified model can be specified to model temporal data $\mathcal{D}=\{\mathbf{y}_i,t_i\}_{i=1}^N$ in a straightforward way. In a few lines, see also Figure~\ref{graphicalmodel}, we write
\begin{align}
	&\mathbf{x}_t = \mathbf{x}_0 + \int_{0}^{t}\mu(\mathbf{x}_s)ds+\int_{0}^{t}\sqrt{\mathbf{\Sigma}(\mathbf{x}_s)}dB_s,\label{lsde-dyn1}\\
	&f(\cdot)|\mathbf{\Sigma}(\cdot),\mathcal{D}\sim\mathcal{GP}(\mu(\cdot),\mathbf{\Sigma}(\cdot)),\\
	&\mathbf{\Sigma}(\cdot)\sim \mathcal{WP}(\cdot|\mathcal{D}),\label{lsde-dyn2}\\
	&p(\mathbf{y}_t|\mathbf{x}_t)=\mathcal{N}(g(\mathbf{x}_t),\mathbf{A\Sigma}(\mathbf{x}_t)\mathbf{A}^\top +\mathbf{\Lambda}).\label{lsde-llh}
\end{align}

If $g$ is not the identity mapping, we can define a \emph{latent dynamical} model. Say $g$ is a GP mapping from $\mathbb{R}^D$ to $\mathbb{R}^\eta$. This is similar to GP state space models (GPSSM) where the dynamics, or transitions, are defined $\mathbf{x}_t = f(\mathbf{x}_{t-1})+\mathbf{\epsilon}_x$ and $\mathbf{y}_t=g(\mathbf{y}_t)+\mathbf{\epsilon}_y$, for GPs $f$ and $g$ and some noise variables $\mathbf{\epsilon}_x$ and $\mathbf{\epsilon}_y$, usually Gaussian with zero mean and diagonal covariance matrix~\citep{Deisenroth2012, Eleftheriadis2017a}.

The latent dynamics defined in \eqref{lsde-dyn1}--\eqref{lsde-dyn2} are not restricted to have equi-temporal measurements and model non-diagonal covariance structure both in the latent states $\mathbf{x}$ and in the observed states $\mathbf{y}$ through the matrix $\mathbf{A}$, which is an $\eta\!\times\! D$-matrix. Adding the diagonal $\eta\!\times\!\eta$-matrix $\mathbf{\Lambda}$ is necessary to avoid singularity. Even though $\boldsymbol\Sigma(\cdot)$ is a $D\times D$-matrix, we can still lower-rank approximate with a $\rho$-rank matrix, as described in Section \ref{sec:complex}. The log-likelihood we compute is
\begin{align}\label{t-llh}
\begin{split}
	\log p(\mathbf{y}_t|\mathbf{g}_t,\mathbf{\Sigma}(\mathbf{x}_t))&=\frac{\eta}{2}\log(2\pi)-\log(\det(\mathbf{B}))\\
&\quad-\frac{1}{2}(\mathbf{y}_t-\mathbf{g}_t)^\top \mathbf{B}^{-1}(\mathbf{y}_t-\mathbf{g}_t),
\end{split}
\end{align}
where $\mathbf{B}:=\mathbf{A\Sigma}(\mathbf{x}_t)\mathbf{A}^\top +\mathbf{\Lambda}$.
As a consequence of the matrix-determinant lemma and the Woodbury identity, we can evaluate the likelihood cheaply, because of $\mathbf{B}$'s structure. The ELBO that we optimise during training is similar to \eqref{ELBO}, only the likelihood term is different: it is swapped for a variational expectation over \eqref{t-llh}. We assume independence between all temporal observations, i.e. $p(\mathcal{D})=\prod_{i=1}^N p(\{\mathbf{y}_i,t_i\})$.

\section{Experiments}\label{sec:experiments}
We evaluate the presented model in both regression and a dynamical setup. In both instances, we use baselines that are similar to our model to easier distinguish the influence the diffusion has on the experiments. We evaluate on a well-studied regression benchmark and on a higher-dimensional dynamical dataset.
\subsection{Regression}
	\begin{figure*}[t]
		\centering
		\includegraphics[width=\hsize]{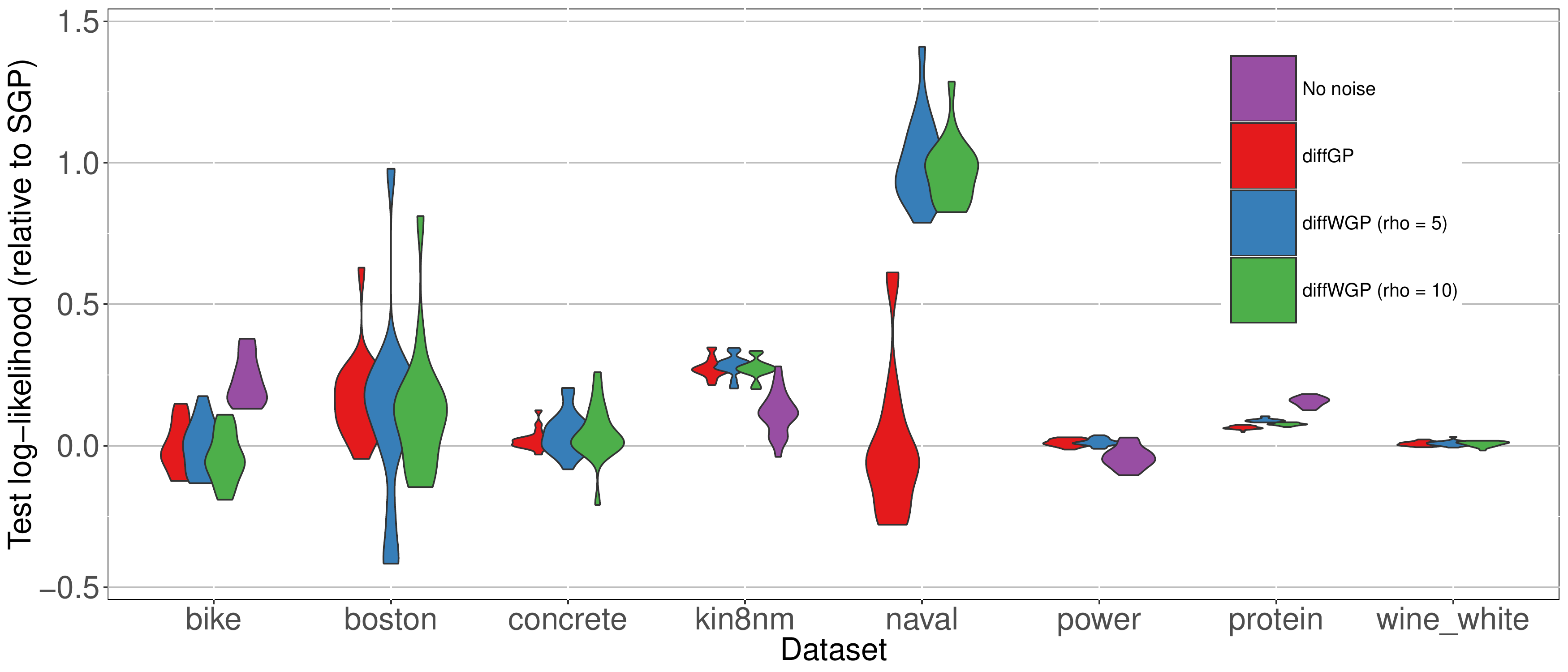}
		\vspace{-2mm}
		\caption{Test-set log-likelihood values on eight UCI regression datasets. The violin plots show the test-set log (likelihood-ratio) of baseline diffusion models with respect to the SGP baseline. Values greater than $0$ indicate an improvement over SGP. Key findings are that No noise can  overfit heavily (boston, concrete, naval), and diffWGP performs best on most datasets. The figure has been cut for readability---this explain why occasionally purple violins are missing.}\label{uciplot}
	\end{figure*}
	We compare our model, which we will dub \emph{Wishart-priored GP flow} (\emph{diffWGP}), to three baseline models in order to shed light on some properties of the diffWGP.
	
	\paragraph*{GP flows} Reproducing the model from \citet{hegde2019deep}  will give indications, if it is possible to increase overall performance by modelling the randomness in the flow. This model has a diagonal matrix $\mathbf{\Sigma}$ with entries determined solely by the chosen covariance function. We will refer to this model with \emph{diffGP}.
	
	\paragraph*{No noise flows} 
	We also evaluate the model, where $\mathbf{\Sigma}=\mathbf{0}$, i.e. the situation where the flow is \emph{deterministic}. The remaining part of the flow is still as in \eqref{flow-mean} to make fair comparisons. All the relevant KL-terms are removed from the ELBO \eqref{ELBO}. We refer to this as \emph{No noise}.
	
	\paragraph*{Sparse GPs} Also in the variational setup we shall compare to vanilla sparse GPs, i.e. the model introduced by \citet{titsias}. We will refer to this as \emph{SGP}.
	
\subsubsection{Experimental Setup}
	In all experiments, we choose $100$ inducing points for the variational distributions, all of which are Gaussians. All models are trained for $50000$ iterations with a mini-batch size of $2000$, or the number of samples in the data if smaller. In all instances, the first $10000$ iterations are \emph{warm-starting} the final layer GP $g$, keeping all other parameters fixed. We use the Adam-optimiser with a step-size of $0.01$. After this all flows (this excludes SGP) are initialised with a constant mean $0$ and covariance functions chosen as RBF with automatic relevance determination (ARD), initialised with tiny signal noise to ensure $\mathbf{x}_0\approx \mathbf{x}_T$. The time variable $T$ is always $1$.
	
	The remaining 40000 iterations (SGP excluded) are updating again with Adam with a more cautious step-size of $0.001$. For the diffWGP, the first 4000 of these are warm-starting the KL-terms associated with the flow to speed up convergence. Note that this model fits more parameters than the baseline models. For the diffWGP, we update the ELBO
	\begin{align}
	\begin{split}
	&\mathbb{E}_{q(g)}[\log p(y|g)]-\text{KL}\big(q(\mathbf{u}_g)\|p(\mathbf{u}_g)\big)\\
	& \quad-c^2\text{KL}\big(q(\mathbf{u}_f)\|p(\mathbf{u}_f)\big)\! -\! c\text{KL}\big(q(\mathbf{u}_\Sigma)\|p(\mathbf{u}_\Sigma)\big),
	\end{split}
	\end{align}
	where $c=\min(1,\frac{iteration}{4000})$, i.e. we warm-start the regularising KL-terms.
	\subsubsection{UCI Regression Benchmark}
	\begin{table}
	\centering
	\resizebox{\linewidth}{!}{
		\begin{tabular}{l|cc}
			& \multicolumn{1}{l}{\textit{diffGP vs. SGP}} & \multicolumn{1}{l}{\textit{diffWGP vs. diffGP}} \\ \hline
			\textsc{Bike} (14)        & 0.8695                                      & 0.2262                                          \\
			\textsc{Boston} (13)    & \textbf{\textless{}0.0001}                  & 0.9867                                          \\
			\textsc{Concrete} (8)    & \textbf{0.0042}                             & \textbf{0.0348}                                 \\
			\textsc{kin8nm} (8)   & \textbf{\textless{}0.0001}                  & \textbf{0.0164}                                 \\
			\textsc{Naval} (26)      & 0.8695                                      & \textbf{\textless{}0.0001}                      \\
			\textsc{Power}  (4)     & \textbf{\textless{}0.0001}                  & 0.1387                                          \\
			\textsc{Protein} (9)    & \textbf{\textless{}0.0001}                  & \textbf{\textless{}0.0001}                      \\
			\textsc{Wine\_white} (11) & \textbf{0.0003}                             & 0.3238                                         
		\end{tabular}
		}
		\caption{Wilcoxons paired signed rank-test. Listed are the p-values of the hypothesis of equal median versus alternative that location shift is negative. Bold highlights the significant ones at a $0.05$ confidence level. In parenthesis are the input dimensionality of the datasets. Results are for $\rho=5$.}
		\label{table-p}
	\end{table}
		\begin{figure*}[t]
\centering
        \subfigure[The log-likelihood of \emph{forecasted} measurements up to 48 hours. The bold lines mark the average log-likelihood at a given hour based on 50 simulations. The associated shaded areas span twice the standard error.]{
		\includegraphics[width=0.75\hsize]{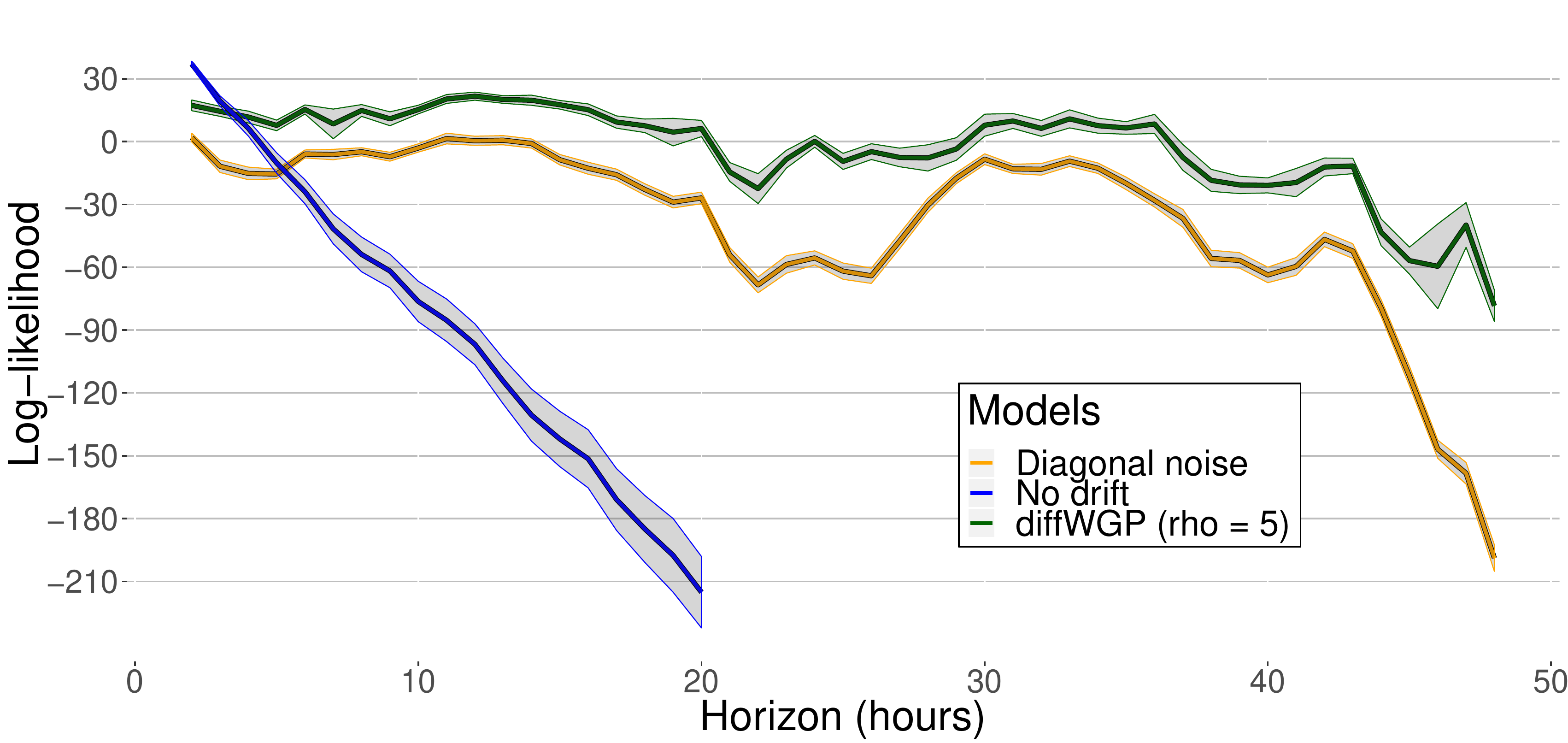}
		%\caption{The log-likelihood of \emph{forecasted} measurements up to 48 hours. The bold lines mark the average log-likelihood at a given hour based on 50 simulations. The associated shaded areas span two times the standard error.}
		\label{autoreg-plot}
		}
		\subfigure[The density (colour) of the 48-hour horizon predictions of temperature measurement in Tiantan ($x$-axis) and Dongsi ($y$-axis). These locations are within a few kilometres of each other. \textit{Left}: diagonal noise case; \textit{Right}: Wishart noise. The Wishart detects a correlation between these two temperature measurements, as we would expect for such nearby locations.]{
		\includegraphics[width=0.75\hsize]{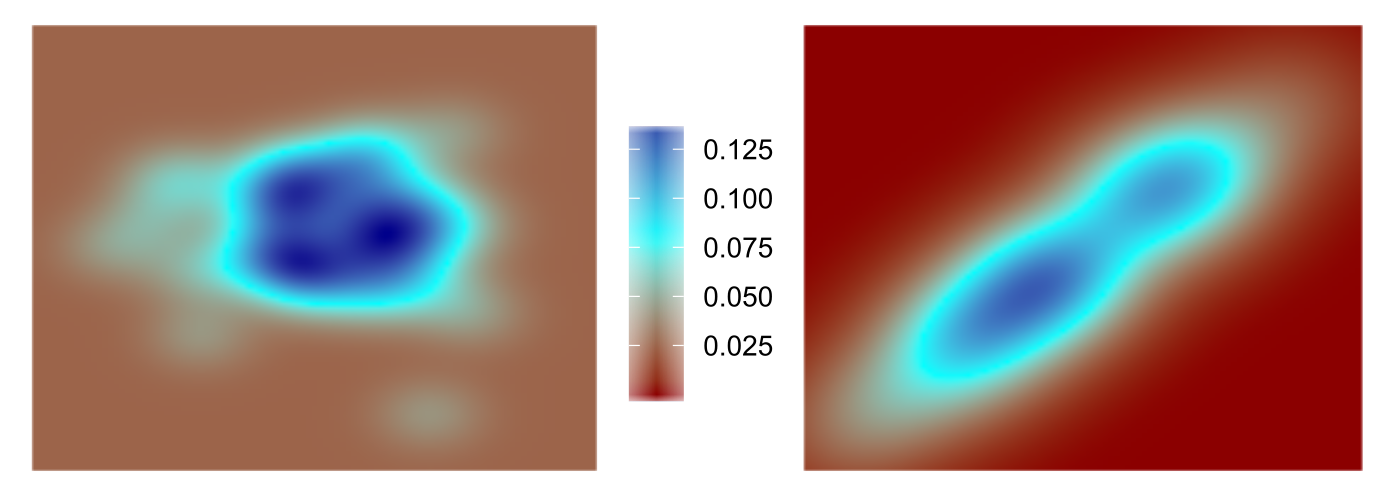}
		\label{what-wishart-learn}
		}
		\caption{(a): The performance of predictions plotted over the forecast horizon. (b): The joint development of two temperature measurements over the forecasted time-horizon for two different models.}
\end{figure*} 
	Figure \ref{uciplot} shows the results on eight UCI benchmark datasets over 20 train-test splits (90/10). On the $y$-axis we see the distribution of the test-set log-likelihood subtracted by the SGP log-likelihood on the same split. Values greater than $0$ are improvements over the baseline SGP. 
	An analogous plot with RMSE is supplied in the supplementary material. In Table \ref{table-p}, we use Wilcoxon's paired rank test to evaluate whether the more advanced models perform better. 

	Key observations are: \emph{not} having noise in the flow (No noise) seem to lead to overfitting, except in two cases, where a more expressive model is preferred. In one of these cases (protein) Wishart modelling improves both the RMSE and the log-likelihood. In one case (boston),  overfitting was absurdly large: on this dataset we were not able to reproduce the results from \citet{hegde2019deep} either. In four cases (concrete, kin8nm, power, wine\_white), \textit{No noise} overfitted mildly. In two of these cases, \emph{diffWGP} improved over \emph{diffGP}. The two cases, where no improvement is significant, are simple cases, wine\_white and power, which are almost linear or low-dimensional. On the naval dataset, the \emph{No noise} model could not run due to numerical issues. Here \emph{diffWGP} outperforms \emph{diffGP} in the log-likelihood. We conjecture this is because of the high dimensionality and the fact that almost no observation noise is present. We found no substantial influence of the parameter $\rho$; if any then it actually seems to prefer lower-rank approximations. This emphasises that training Wishart processes is difficult, and further research in this area is needed.
	
	\subsection{Auto-regressive modelling of air quality}
	We evaluate our dynamical model on atmospheric air-quality data from Beijing \citep{china-data}. We pre-processed the data for three locations in the city (Shunyi, Tiantan, Dongsi), which each have hourly observation of ten features over the period of 2014--2016. Explicitly, the ten features are: the concentration of PM2.5, PM10, SO2, NO2, CO, O3, the temperature and dew point temperature, air pressure and amount of precipitation.\footnote{The full data set is available from \href{https://archive.ics.uci.edu/ml/datasets/Beijing+Multi-Site+Air-Quality+Data}{here}.}
	
	We use the first two years of this dataset for training and aim to forecast into the first 48 hours of 2016. Including the variables year, month, day and hour, we have in total 34 features for the three cities and 17520 temporal observations for training. Missing values were linearly interpolated. All features were standardised.
	
	To analyse properties of our proposed model, we perform an ablation study with the following models:
	
	\textbf{diffWGP} The model proposed in the paper to model the diffusion with Wisharts.
	
	\textbf{Diagonal noise} The drift term remains as in the diffWGP model, but the diffusion is restricted to diagonal, i.e. correlated diffusion cannot be modelled. This becomes the model
	\begin{equation}
	    \mathbf{x}_t = \mathbf{x}_s +\mu(\mathbf{x}_s)(t-s) + \sqrt{\mathbf{\Lambda}(t-s)}\mathbf{\epsilon}_t, \quad \mathbf{\epsilon}\sim\mathcal{N}(\mathbf 0,\mathbf{I}).
	\end{equation} 
	
	\textbf{No drift} The drift is constantly zero, and the diffusion is modelled by a Wishart, which results in the model 
	\begin{equation}
	\mathbf{x}_t = \mathbf{x}_s+\sqrt{\big(\mathbf{A\Sigma}(\mathbf{x}_t)\mathbf{A}^\top +\mathbf{\Lambda}\big)(t-s)}\mathbf{\epsilon}_t.
	\end{equation}
	This model is a continuous-time version of the model presented by \citet{wilk_wish}.
    
    In all instances, we train by minibatching shorter sequences, and we use the Adam optimiser \citep{kingma2014adam} with a learning rate 0.01. Due to the large amount of temporal observation compared to small batches we ease off on momentum.

	Figure~\ref{autoreg-plot} shows how the different models \emph{forecast} future observations by reporting the log-likelihood traces of individual models at test time. The figure shows the mean and two times the standard error, which we obtain from 50 simulations. At first, we see that having no drift starts off better, but quickly drops in performance. This is not unexpected, as the data has structure in its evolution.
	The difference between the models with drift, but different diffusions, are more interesting for this dataset. Overall, Wishart diffusions perform best, and it seems to be resilient and take only few and relatively small `dips'.
    
    We expect this dataset to have highly correlated features. The three locations in Beijing are, in distance, close to each other; naturally the different air measurements are similar in their evolution over time. Figure~\ref{what-wishart-learn} illustrates how a model with diagonal noise is incapable of learning this joint development of temperature measurements. Here, the Wishart learns that when the temperature in Dongsi is high, it is also high in Tiantan. This behaviour is seen in many pairs of the features considered, and it suggests \emph{diffWGP} has dynamics moving on a manifold of smaller dimension than if diagonal noise was considered. This supports the hypothesis that \emph{diffWGP} moves as \emph{one} dynamical systems, opposed to 34.

	\section{Conclusion}
	In a non-parametric Bayesian way, we presented a scalable approach to continuous-time learning with high emphasis on correlated process noise. This noise is modelled with a Wishart process, which lets high-dimensional data evolve as a single system, rather than $D$ independent systems. We presented a way to scale this to high dimensions. We found that it is never worse taking the dependence structure in the process noise into account. However, with certain types of data, it can mitigate overfitting effects and improve performance.
	
	Code is publicly available at: \url{https://github.com/JorgensenMart/Wishart-priored-SDE}.

\subsection*{Acknowledgements}
MJ was supported by a research grant (15334) from VILLUM FONDEN.
\bibliographystyle{icml2020}
\bibliography{references}

\begin{thebibliography}{35}
\providecommand{\natexlab}[1]{#1}
\providecommand{\url}[1]{\texttt{#1}}
\expandafter\ifx\csname urlstyle\endcsname\relax
  \providecommand{\doi}[1]{doi: #1}\else
  \providecommand{\doi}{doi: \begingroup \urlstyle{rm}\Url}\fi

\bibitem[{\'A}lvarez \& Lawrence(2011){\'A}lvarez and Lawrence]{alvarez2011}
{\'A}lvarez, M.~A. and Lawrence, N.~D.
\newblock Computationally efficient convolved multiple output {G}aussian
  processes.
\newblock \emph{Journal of Machine Learning Research}, 12:\penalty0 1459--1500,
  2011.

\bibitem[Andreas \& Kandemir(2019)Andreas and Kandemir]{look}
Andreas, L. and Kandemir, M.
\newblock Differential {B}ayesian neural nets.
\newblock \emph{arXiv:1912.00796}, 2019.

\bibitem[Chen et~al.(2018)Chen, Rubanova, Bettencourt, and
  Duvenaud]{chen2018neural}
Chen, T.~Q., Rubanova, Y., Bettencourt, J., and Duvenaud, D.~K.
\newblock Neural ordinary differential equations.
\newblock In \emph{Advances in Neural Information Processing Systems}, 2018.

\bibitem[Damianou \& Lawrence(2013)Damianou and Lawrence]{damianou2013deep}
Damianou, A. and Lawrence, N.~D.
\newblock Deep {G}aussian processes.
\newblock In \emph{Artificial Intelligence and Statistics}, 2013.

\bibitem[Deisenroth et~al.(2012)Deisenroth, Turner, Huber, Hanebeck, and
  Rasmussen]{Deisenroth2012}
Deisenroth, M.~P., Turner, R., Huber, M., Hanebeck, U.~D., and Rasmussen, C.~E.
\newblock Robust filtering and smoothing with {Gauss}ian processes.
\newblock \emph{IEEE Transactions on Automatic Control}, 57\penalty0
  (7):\penalty0 1865--1871, 2012.

\bibitem[Dupont et~al.(2019)Dupont, Doucet, and Teh]{dupont}
Dupont, E., Doucet, A., and Teh, Y.~W.
\newblock Augmented neural {ODE}s.
\newblock In \emph{Advances in Neural Information Processing Systems}, 2019.

\bibitem[Durrett(2018)]{durrett}
Durrett, R.
\newblock \emph{Stochastic calculus: a practical introduction}.
\newblock CRC press, 2018.

\bibitem[Duvenaud et~al.(2014)Duvenaud, Rippel, Adams, and
  Ghahramani]{duvenaud2014avoiding}
Duvenaud, D., Rippel, O., Adams, R., and Ghahramani, Z.
\newblock Avoiding pathologies in very deep networks.
\newblock In \emph{Artificial Intelligence and Statistics}, 2014.

\bibitem[E(2017)]{E2017}
E, W.
\newblock A proposal on machine learning via dynamical systems.
\newblock \emph{Communications in Mathematics and Statistics}, 5\penalty0
  (1):\penalty0 1--11, 2017.

\bibitem[Eleftheriadis et~al.(2017)Eleftheriadis, Nicholson, Deisenroth, and
  Hensman]{Eleftheriadis2017a}
Eleftheriadis, S., Nicholson, T. F.~W., Deisenroth, M.~P., and Hensman, J.
\newblock Identification of {G}aussian process state space models.
\newblock In \emph{Advances in Neural Information Processing Systems}, 2017.

\bibitem[Engle(1982)]{engle1982autoregressive}
Engle, R.~F.
\newblock Autoregressive conditional heteroscedasticity with estimates of the
  variance of {United Kingdom} inflation.
\newblock \emph{Econometrica: Journal of the Econometric Society}, pp.\
  987--1007, 1982.

\bibitem[Haber \& Ruthotto(2017)Haber and Ruthotto]{Haber2017}
Haber, E. and Ruthotto, L.
\newblock Stable architectures for deep neural networks.
\newblock \emph{Inverse Problems}, 34\penalty0 (1):\penalty0 014004, 2017.

\bibitem[He et~al.(2016)He, Zhang, Ren, and Sun]{he2016deep}
He, K., Zhang, X., Ren, S., and Sun, J.
\newblock Deep residual learning for image recognition.
\newblock In \emph{Conference on Computer Vision and Pattern Recognition},
  2016.

\bibitem[Heaukulani \& van~der Wilk(2019)Heaukulani and van~der
  Wilk]{wilk_wish}
Heaukulani, C. and van~der Wilk, M.
\newblock Scalable {B}ayesian dynamic covariance modeling with variational
  {W}ishart and inverse {W}ishart processes.
\newblock In \emph{Advances in Neural Information Processing Systems}, 2019.

\bibitem[Hegde et~al.(2019)Hegde, Heinonen, L{\"a}hdesm{\"a}ki, and
  Kaski]{hegde2019deep}
Hegde, P., Heinonen, M., L{\"a}hdesm{\"a}ki, H., and Kaski, S.
\newblock Deep learning with differential {G}aussian process flows.
\newblock In \emph{Artificial Intelligence and Statistics}, 2019.

\bibitem[Hensman et~al.(2013)Hensman, Fusi, and Lawrence]{bigdata-hensman}
Hensman, J., Fusi, N., and Lawrence, N.~D.
\newblock Gaussian processes for big data.
\newblock In \emph{Uncertainty in Artificial Intelligence}, 2013.

\bibitem[Hensman et~al.(2015)Hensman, Matthews, Filippone, and
  Ghahramani]{hensman2015mcmc}
Hensman, J., Matthews, A. G. d.~G., Filippone, M., and Ghahramani, Z.
\newblock {MCMC} for variationally sparse {G}aussian processes.
\newblock In \emph{Advances in Neural Information Processing Systems}, 2015.

\bibitem[Itô(1946)]{ito}
Itô, K.
\newblock On a stochastic integral equation.
\newblock \emph{Proceedings of the Japan Academy}, 22\penalty0 (2):\penalty0
  32--35, 1946.

\bibitem[Kingma \& Ba(2014)Kingma and Ba]{kingma2014adam}
Kingma, D.~P. and Ba, J.
\newblock Adam: A method for stochastic optimization.
\newblock \emph{arXiv:1412.6980}, 2014.

\bibitem[Kloeden \& Platen(2013)Kloeden and Platen]{kloeden2013numerical}
Kloeden, P.~E. and Platen, E.
\newblock \emph{Numerical Solution of Stochastic Differential Equations},
  volume~23.
\newblock Springer Science \& Business Media, 2013.

\bibitem[Kshirsagar(1959)]{kshirsagar1959bartlett}
Kshirsagar, A.~M.
\newblock Bartlett {D}ecomposition and {W}ishart distribution.
\newblock \emph{The Annals of Mathematical Statistics}, 30\penalty0
  (1):\penalty0 239--241, 1959.

\bibitem[Li et~al.(2020)Li, Wong, Chen, and Duvenaud]{li2020scalable}
Li, X., Wong, T.-K.~L., Chen, R. T.~Q., and Duvenaud, D.
\newblock Scalable gradients for stochastic differential equations.
\newblock In \emph{Artificial Intelligence and Statistics}, 2020.

\bibitem[Liu et~al.(2019)Liu, Xiao, Si, Cao, Kumar, and Hsieh]{liu}
Liu, X., Xiao, T., Si, S., Cao, Q., Kumar, S., and Hsieh, C.-J.
\newblock Neural {SDE}: Stabilizing neural {ODE} networks with stochastic
  noise.
\newblock \emph{arXiv:1906.02355}, 2019.

\bibitem[Qui\~{n}onero Candela \& Rasmussen(2005)Qui\~{n}onero Candela and
  Rasmussen]{sparse-unifying}
Qui\~{n}onero Candela, J. and Rasmussen, C.~E.
\newblock A unifying view of sparse approximate {G}aussian process regression.
\newblock \emph{Journal of Machine Learning Research}, 6:\penalty0 1939--1959,
  2005.

\bibitem[Rasmussen \& Williams(2006)Rasmussen and Williams]{rasmussen:book}
Rasmussen, C.~E. and Williams, C.
\newblock \emph{{Gaussian Processes for Machine Learning}}.
\newblock MIT Press, 2006.

\bibitem[Salimans \& Knowles(2013)Salimans and Knowles]{salimans2013fixed}
Salimans, T. and Knowles, D.~A.
\newblock Fixed-form variational posterior approximation through stochastic
  linear regression.
\newblock \emph{Bayesian Analysis}, 8\penalty0 (4):\penalty0 837--882, 2013.

\bibitem[Salimbeni \& Deisenroth(2017)Salimbeni and
  Deisenroth]{salimbeni2017doubly}
Salimbeni, H. and Deisenroth, M.~P.
\newblock Doubly stochastic variational inference for deep {G}aussian
  processes.
\newblock In \emph{Advances in Neural Information Processing Systems}, 2017.

\bibitem[Seeger et~al.(2005)Seeger, Teh, and Jordan]{SLFM}
Seeger, M., Teh, Y.-W., and Jordan, M.
\newblock Semiparametric latent factor models.
\newblock Technical report, 2005.

\bibitem[Snelson \& Ghahramani(2006)Snelson and Ghahramani]{snelson2006sparse}
Snelson, E. and Ghahramani, Z.
\newblock Sparse {G}aussian processes using pseudo-inputs.
\newblock In \emph{Advances in Neural Information Processing Systems}, 2006.

\bibitem[Titsias(2009)]{titsias}
Titsias, M.
\newblock Variational learning of inducing variables in sparse {G}aussian
  processes.
\newblock In \emph{Artificial Intelligence and Statistics}, 2009.

\bibitem[Twomey et~al.(2019)Twomey, Koz{\l}owski, and
  Santos-Rodr{\'\i}guez]{twomey}
Twomey, N., Koz{\l}owski, M., and Santos-Rodr{\'\i}guez, R.
\newblock Neural {ODE}s with stochastic vector field mixtures.
\newblock \emph{arXiv:1905.09905}, 2019.

\bibitem[Tzen \& Raginsky(2019)Tzen and Raginsky]{tzen}
Tzen, B. and Raginsky, M.
\newblock Neural stochastic differential equations: deep latent {G}aussian
  models in the diffusion limit.
\newblock \emph{arXiv:1905.09883}, 2019.

\bibitem[Wilson \& Ghahramani(2010)Wilson and Ghahramani]{GWP}
Wilson, A.~G. and Ghahramani, Z.
\newblock Generalised {W}ishart processes.
\newblock \emph{arXiv:1101.0240}, 2010.

\bibitem[Wu et~al.(2014)Wu, Hern\'{a}ndez-Lobato, and Ghahramani]{gp_vol_wu}
Wu, Y., Hern\'{a}ndez-Lobato, J.~M., and Ghahramani, Z.
\newblock Gaussian process volatility model.
\newblock In \emph{Advances in Neural Information Processing Systems}. 2014.

\bibitem[Zhang et~al.(2017)Zhang, Guo, Dong, He, Xu, and Chen]{china-data}
Zhang, S., Guo, B., Dong, A., He, J., Xu, Z., and Chen, S.~X.
\newblock Cautionary tales on air-quality improvement in {B}eijing.
\newblock \emph{Proceedings of the Royal Society A: Mathematical, Physical and
  Engineering Sciences}, 473, 2017.

\end{thebibliography}
\newpage
\onecolumn
\section*{Supplementary material}
\begin{figure}[h]
	\centering
	\includegraphics[width=\textwidth]{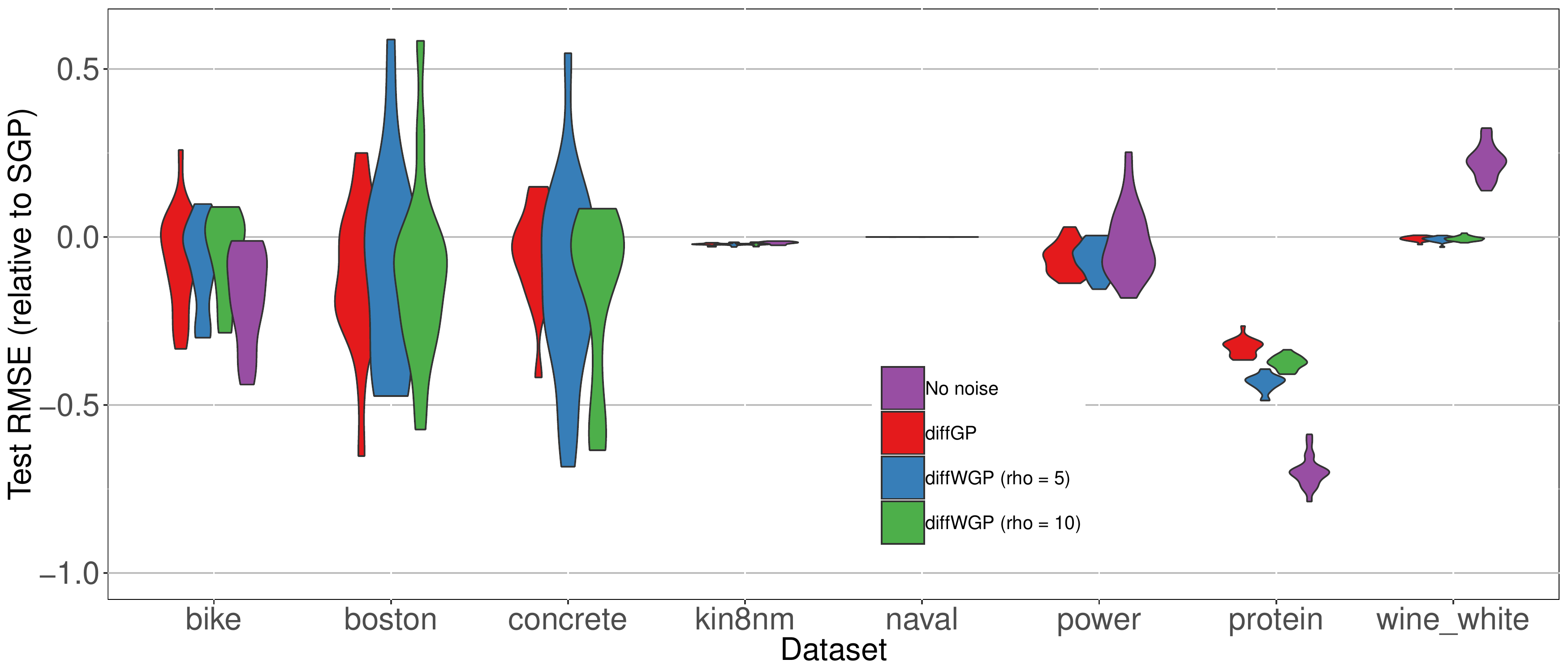}
	\caption{Test-set RMSE values on eight UCI regression datasets. The violin plots show the test-set RMSE difference of baseline diffusion models with respect to the SGP baseline. Values \emph{less} than 0 indicate an improvement over SGP. We remark that this Figure has been cut for readability---this explains why occasionally purple violins are missing.}
	\label{uciplot_rmse}
\end{figure}

A plot of the test-set RMSE, relative to SGP, is provided in Figure \ref{uciplot_rmse}. Further, we have included the actual results for reproducability; these can be seen in Table \ref{table}.
\begin{table}[h]
\resizebox{\textwidth}{!}{
\begin{tabular}{clrrrr}
\multicolumn{2}{c}{Dataset}                                                           & \multicolumn{1}{c}{bike}                     & \multicolumn{1}{c}{boston}                       & \multicolumn{1}{c}{concrete}                  & \multicolumn{1}{c}{kin8nm}                  \\
\multicolumn{2}{c}{Dimension}                                                         & \multicolumn{1}{c}{14}                       & \multicolumn{1}{c}{13}                           & \multicolumn{1}{c}{8}                         & \multicolumn{1}{c}{8}                       \\ \hline
                                          & \cellcolor[HTML]{C0C0C0}No noise          & \cellcolor[HTML]{C0C0C0}$-0.4169\pm 0.0235$  & \cellcolor[HTML]{C0C0C0}$-1442.5453\pm 512.4655$ & \cellcolor[HTML]{C0C0C0}$-25.4477\pm 12.9521$ & \cellcolor[HTML]{C0C0C0}$1.1458\pm 0.0696$  \\
                                          & diffGP                                    & $-0.6842\pm 0.1143$                          & $-2.7586\pm 0.7301$                              & $-3.0401\pm 0.1460$                           & $1.3046\pm 0.0302$                          \\
                                          & \cellcolor[HTML]{C0C0C0}diffWGP (rho = 5) & \cellcolor[HTML]{C0C0C0}$-0.6569\pm 0.0381$  & \cellcolor[HTML]{C0C0C0}$-3.1014\pm 0.8984$      & \cellcolor[HTML]{C0C0C0}$-3.0164\pm 0.1352$   & \cellcolor[HTML]{C0C0C0}$1.3123\pm 0.0305$  \\
\multirow{-4}{*}{Test-set log-likelihood} & diffWGP (rho = 10)                        & $-0.7171\pm 0.1621$                          & $-2.9540\pm 0.8506$                              & $-3.0141\pm 0.1559$                           & $1.3021\pm 0.0304$                          \\ \hline
                                          & \cellcolor[HTML]{C0C0C0}No noise          & \cellcolor[HTML]{C0C0C0}$0.0706\pm 0.0305$   & \cellcolor[HTML]{C0C0C0}$4.7823\pm 1.0100$       & \cellcolor[HTML]{C0C0C0}$7.5674\pm 1.3031$    & \cellcolor[HTML]{C0C0C0}$0.0688\pm 0.0026$  \\
                                          & diffGP                                    & $0.1845\pm 0.0507$                           & $2.9628\pm 0.6991$                               & $4.9913\pm 0.6657$                            & $0.0646\pm 0.0021$                          \\
                                          & \cellcolor[HTML]{C0C0C0}diffWGP (rho = 5) & \cellcolor[HTML]{C0C0C0}$0.1737\pm 0.0254$   & \cellcolor[HTML]{C0C0C0}$3.1713\pm 0.8296$       & \cellcolor[HTML]{C0C0C0}$4.9086\pm 0.6442$    & \cellcolor[HTML]{C0C0C0}$0.0642\pm 0.0023$  \\
\multirow{-4}{*}{Test-set RMSE}           & diffWGP (rho = 10)                        & $0.1935\pm 0.0919$                           & $3.0800\pm 0.8009$                               & $4.8708\pm 0.7079$                            & $0.0649\pm 0.0024$                          \\ \hline
\multicolumn{2}{c}{Dataset}                                                           & \multicolumn{1}{c}{naval}                    & \multicolumn{1}{c}{power}                        & \multicolumn{1}{c}{protein}                   & \multicolumn{1}{c}{wine\_white}             \\
\multicolumn{2}{c}{Dimension}                                                         & \multicolumn{1}{c}{26}                       & \multicolumn{1}{c}{4}                            & \multicolumn{1}{c}{9}                         & \multicolumn{1}{c}{11}                      \\ \hline
                                          & \cellcolor[HTML]{C0C0C0}No noise          & \cellcolor[HTML]{C0C0C0}NA                   & \cellcolor[HTML]{C0C0C0}$-2.8269\pm 0.0618$      & \cellcolor[HTML]{C0C0C0}$-2.7678\pm 0.0187$   & \cellcolor[HTML]{C0C0C0}$-7.2889\pm 0.9605$ \\
                                          & diffGP                                    & $7.2256\pm 0.0949$                           & $-2.7753\pm 0.0402$                              & $-2.8599\pm 0.0109$                           & $-1.0616\pm 0.0448$                         \\
                                          & \cellcolor[HTML]{C0C0C0}diffWGP (rho = 5) & \cellcolor[HTML]{C0C0C0}$8.2736\pm 0.0677$   & \cellcolor[HTML]{C0C0C0}$-2.7736\pm 0.0413$      & \cellcolor[HTML]{C0C0C0}$-2.8342\pm 0.0114$   & \cellcolor[HTML]{C0C0C0}$-1.0603\pm 0.0443$ \\
\multirow{-4}{*}{Test-set log-likelihood} & diffWGP (rho = 10)                        & $8.2696\pm 0.0575$                           & NA                                               & $-2.8480\pm 0.0103$                           & $-1.0620\pm 0.0448$                         \\ \hline
                                          & \cellcolor[HTML]{C0C0C0}No noise          & \cellcolor[HTML]{C0C0C0}NA                   & \cellcolor[HTML]{C0C0C0}$3.8966\pm 0.1674$       & \cellcolor[HTML]{C0C0C0}$3.8029\pm 0.0552$    & \cellcolor[HTML]{C0C0C0}$0.9298\pm 0.0543$  \\
                                          & diffGP                                    & $<0.0001\pm <0.0001$                         & $3.8497\pm \ 0.1496$                              & $4.1717\pm 0.0479$                            & $0.7001\pm 0.0295$                          \\
                                          & \cellcolor[HTML]{C0C0C0}diffWGP (rho = 5) & \cellcolor[HTML]{C0C0C0}$<0.0001\pm <0.0001$ & \cellcolor[HTML]{C0C0C0}$3.8492\pm 0.1502$       & \cellcolor[HTML]{C0C0C0}$4.0677\pm 0.0484$    & \cellcolor[HTML]{C0C0C0}$0.6980\pm 0.0296$  \\
\multirow{-4}{*}{Test-set RMSE}           & diffWGP (rho = 10)                        & $<0.0001\pm <0.0001$                         & NA                                               & $4.1267\pm 0.0441$                            & $0.7003\pm 0.0296$                         
\end{tabular}
}
	\caption{Test set log-likelihood and RMSE on 8 UCI benchmark datasets. Mean and standard deviations over 20 splits.}
	\label{table}
\end{table}

\end{document}